\documentclass[runningheads]{llncs}
\usepackage[T1]{fontenc}
\usepackage{graphicx}
\usepackage{booktabs}
\usepackage{multirow}
\usepackage{xcolor}
\usepackage{float}

\usepackage{appendix}

\usepackage{footnote}
\usepackage{hyperref}
\usepackage{orcidlink} % Support for ORCID icon
\usepackage{marvosym}

\usepackage{multicol}               % 合并多列
\usepackage{utfsym}
\usepackage{colortbl}  %彩色表格需要加载的宏包
\usepackage{xcolor}
\usepackage{pifont}
\usepackage{bm}
\usepackage{amsmath}
\usepackage{amssymb}
\usepackage{mathrsfs}
\usepackage{enumitem}
\usepackage{mdframed}
\usepackage{mathtools}
\usepackage{amsfonts}       % blackboard math symbols

\definecolor{turquoise}{cmyk}{0.65,0,0.1,0.3}
\definecolor{purple}{rgb}{0.65,0,0.65}
\definecolor{dark_green}{rgb}{0, 0.5, 0}
\definecolor{orange}{rgb}{0.8, 0.6, 0.2}
\definecolor{red}{rgb}{0.8, 0.2, 0.2}
\definecolor{darkred}{rgb}{0.6, 0.1, 0.05}
\definecolor{blueish}{rgb}{0.0, 0.3, .6}
\definecolor{light_gray}{rgb}{0.7, 0.7, .7}
\definecolor{pink}{rgb}{1, 0, 1}
\definecolor{greyblue}{rgb}{0.25, 0.25, 1}
\definecolor{igray}{gray}{.9}
\definecolor{pink}{rgb}{1, 0, 1}
\definecolor{pinkishred}{rgb}{0.9647058823529412, 0.6, 0.8196078431372549}
\definecolor{ForestGreen}{RGB}{21,155,82}
\definecolor{brilliantrose}{rgb}{1.0, 0.33, 0.64}
\definecolor{Gray}{gray}{0.9}

\definecolor{citecolor}{HTML}{0071bc} 
\definecolor{SeaGreen4}{RGB}{0,205,102} 
\definecolor{SlateBlue}{RGB}{106,90,205} 
\definecolor{DarkRed}{RGB}{178,34,34} 

\definecolor{CommonR}{RGB}{255,0,0}
\definecolor{R1color}{RGB}{139,54,140}
\definecolor{R2color}{RGB}{244,157,196}
\definecolor{R3color}{RGB}{237,34,104}
\definecolor{R4color}{RGB}{168,2,7}

\newcommand{\yes}{
    \textcolor{SeaGreen4}{\ding{51}}
}
\newcommand{\no}{
    \textcolor{DarkRed}{\ding{55}}
}

\makeatletter
\def\thanks#1{\protected@xdef\@thanks{\@thanks\protect\footnotetext{#1}}}
\makeatother

\def\etal{{\em et al.}}

\begin{document}
\title{Dynamic Prompt Adjustment for Multi-Label Class-Incremental Learning}

\author{Haifeng Zhao\inst{1,2}\orcidlink{0000-0002-5300-0683} \and
Yuguang Jin\inst{1,2}\orcidlink{0009-0005-7700-0054} \and
Leilei Ma\textsuperscript{\Letter}\thanks{\textsuperscript{\Letter} Corresponding Author}\inst{1,2}\orcidlink{0000-0001-8681-0765}}
% %
% \textsuperscript{\Letter}\thanks{\textsuperscript{\Letter} Corresponding Author}
\authorrunning{Zhao et al.}
\institute{ {
Anhui Provincial Key Laboratory of Multimodal Cognitive Computation, Anhui University, Hefei, Anhui, China \\
\and
School of Computer Science and Technology, Anhui University, Hefei, Anhui, China 
\\
\email{\{senith,xiaomylei\}@163.com}
} 
}
\maketitle

\begin{abstract}
Significant advancements have been made in \textit{single}-label incremental learning (SLCIL), yet the more practical and challenging \textit{multi}-label class-incremental learning (MLCIL) remains understudied. 
Recently, visual language models such as CLIP have achieved good results in classification tasks. However, directly using CLIP to solve MLCIL issue can lead to catastrophic forgetting.
To tackle this issue, we integrate an improved data replay mechanism and prompt loss to curb knowledge forgetting.
Specifically, our model enhances the prompt information to better adapt to multi-label classification tasks and employs confidence-based replay strategy to select representative samples. 
Moreover, the prompt loss significantly reduces the model's forgetting of previous knowledge. 
Experimental results demonstrate that our method has substantially improved the performance of MLCIL tasks across multiple benchmark datasets, validating its effectiveness.
\vspace{-0.5em}
\end{abstract}
\section{Introduction}
Class-Incremental Learning (CIL)~\cite{rebuffi2017icarl,van2018generative} focuses on gradually introducing new categories during the training process while maintaining the ability to recognize old categories. This learning approach simulates the ever-changing environment of the real world, where new object categories might appear at any time.

\begin{figure}[t]
\centering
\setlength{\abovecaptionskip}{0.cm}
\includegraphics[width=1.0\textwidth]{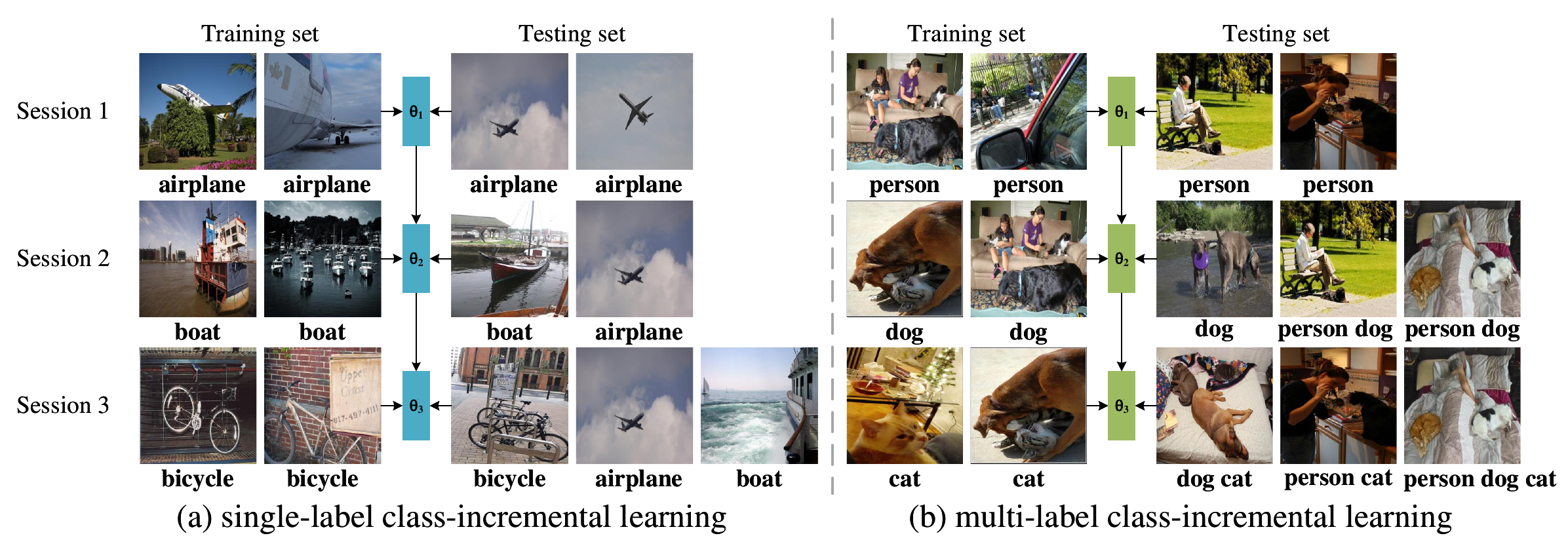}
\caption{Illustration of incremental learning for both single-label and multi-label classification tasks.  It is assumed that three classes are to be learned, with models $\theta_{1}$, $\mathbf{\theta}_{2}$, and $\mathbf{\theta}_{3}$ being trained in consecutive sessions.
} 
\label{figure1}
\vspace{-2.0em} 
\end{figure}

Current research in CIL predominantly addresses the single-label classification challenge~\cite{buzzega2dark020,douillard2020podnet,wang2022learning}, assuming that each image contains single object, as depicted in Fig.~\ref{figure1}(a). However, real-world scenarios often involve multiple objects per image, highlighting the necessity for Multi-Label Class-Incremental Learning (MLCIL). In MLCIL, the model learns with limited class information during each session while managing the absence of labels for previously learned classes.  
As shown in Fig.~\ref{figure1}(b), during the training phase,
image containing {\ttfamily{\{person;cat;dog\}}} is labeled only for {\ttfamily{person}} in the first session. In session 2, this image is re-labeled to include {\ttfamily{dog}}, with {\ttfamily{person}} now classified as negative
class. However, the model is still expected to recognize the {\ttfamily{person}} class in test images. This situation aggravates the forgetting of knowledge in MLCIL. 
Some SLCIL methods~\cite{buzzega2dark020,yan2021dynamically} dynamically expand the model's capacity to handle an increasing number of categories, which may introduce biases, particularly with uneven sample distribution between new and old categories. iCaRL~\cite{rebuffi2017icarl}, for example, selects representative classes for subsequent training but relies on global average pooling, which favors larger objects and may cause sampling imbalance.

To address aforementioned challenge, we intruduce a novel method, 
which consists of two key modules: the Incremental-Context Prompt (ICP) and the Selective Confidence Cluster Replay (SCCR). The ICP employs dual-context prompt mechanism to reduce bias, ensuring that the model remains balanced across different classes. The SCCR utilizes clustering and model confidence to identify and replay significant samples, effectively combating knowledge forgetting. Additionally, we introduce the Textual Prompt Consistency Loss to maintain consistency in textual prompts. Our approach has been validated on the MS COCO and PASCAL VOC datasets, demonstrating its effectiveness in MLCIL.
Overall, our contributions are summarized as follows: (\textbf{1}) We have utilized a new MLCIL framework based on text prompts. As far as we know, this is the first to employ image-text matching to solve the MLCIL problem; (\textbf{2}) We introduce ICP for incremental learning and SCCR for selecting representative samples to alleviate knowledge forgetting; (\textbf{3}) Extensive experiments demonstrate that our method achieves competitive results in addressing the MLCIL problem.

\section{Related Work}
\vspace{-0.5em}
\subsection{Class-Incremental Learning}
CIL has made significant progress. Approaches in CIL mainly include regularization-based, rehearsal-based, and architecture-based methods.
\textbf{Regularization-based} 
methods~\cite{li2017learning,schwarz2018progress} introduce regularization term into the loss function to constrain the variation of model parameters. For example, EWC~\cite{kirkpatrick2017overcoming} uses a Fisher information matrix to mitigate changes in important parameters associated with old tasks. 
\textbf{Rehearsal-based}  
methods~\cite{buzzega2dark020,douillard2020podnet,riemer2018learning,shin2017continual,tao2020topology} 
prevent catastrophic forgetting by replaying a small number of samples or features from old tasks.
 Xiang \etal~\cite{xiang2019incremental} generates pseudo-features of old categories using generative adversarial networks
which reduces memory usage.
\textbf{Architectural-based} method~\cite{de2024less,du2022agcn} provides independent parameters for each task.
PackNet~\cite{mallya2018packnet} proposes to isolate the old and new task parameters for knowledge retention. 
However, this approach can lead to substantial increase in the total number of parameters.
\vspace{-1.0em}
\subsection{Prompt Tuning for Incremental Learning}
Prompt tuning~\cite{jia2022visual} enables models to achieve high
efficiency on downstream tasks with minimal parameter adjustments. Key innovations in this area include L2P's~\cite{wang2022learning} introduction of prompt pools; DualPrompt's~\cite{wang2022dualprompt} distinction between general and specific knowledge through G-Prompt and E-Prompt; and AttriCLIP's~\cite{wang2023attriclip} use of textual prompts to refine model understanding. However, these methods are primarily designed for SLCIL. We tackle MLCIL by integrating prompt learning techniques inspired by CoOp~\cite{zhou2022learning} and leveraging the strengths of CLIP~\cite{radford2021learning}.
\vspace{-1.0em} 
\section{Methodology}
\subsection{Framework}
\begin{figure}[t]
\centering
\setlength{\abovecaptionskip}{0.cm}
\includegraphics[width=1.0\textwidth]{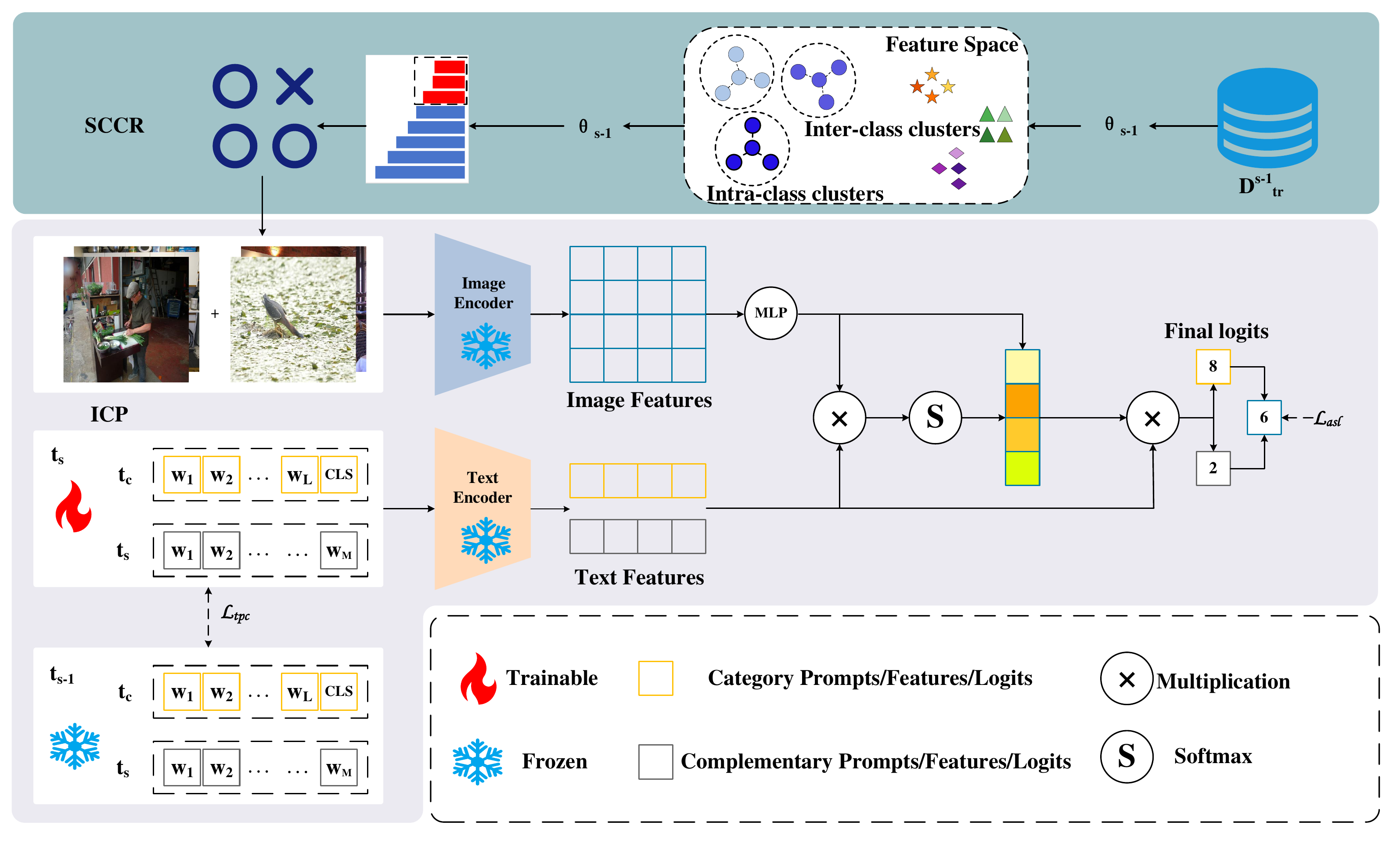}
\caption{
Overview of our model. 
}
\label{figure2}
\vspace{-2.0em} 
\end{figure}

As shown in Fig.~\ref{figure2}, our model mainly comprises two key components: 
\textit{Incremental Context Prompting} (ICP)  and \textit{Selective Confidence Cluster Replay} (SCCR) module. During session 
$S^n$, the SCCR module selects samples from the previous $n-1$ sessions, which are combined with the current training set $\mathcal{D}_{tr}^n$ for joint training.  Simultaneously, ICP learns category-specific prompts ($\mathcal{t}_{c}$) and supplementary
% \vspace{-0.5em}
context prompts ($\mathcal{t}_{s}$). These prompt features are aligned with image
features to produce classification results. 
\vspace{-1.0em} 
\subsection{Incremental Context Prompting Learner}
Inspired by CoOp~\cite{zhou2022learning}, we introduce ICP for MLCIL. Within session $n$, 
we devise learnable prompts $\bm{p}_c$ encompassing all existing categories:
\begin{equation}\label{eqn-1} 
  % \boldsymbol{t}_c=\left[ \omega_1,\omega_2, \dots, \omega_L, {\rm CLS}_{j}\right]
  \bm{t}_c=\left[ \omega_1,\omega_2, \dots, \omega_L, {\rm CLS}_{j}\right]~\text{,}
\end{equation}
where $\omega_{i}$ denoting the learnable tokens and $[{\rm CLS}_{j}]$ representing the word embedding specific to the $j$-th category. The length of learnable tokens is L.
% Contextual information surrounding the categories, including background elements, has a significant impact on classification results.
Motivated by \cite{sun2022dualcoop}, 
we integrate contextual prompt $\bm{t}_{s}$ to complement $\bm{t}_{c}$. 
The form is as follows:
\begin{equation}\label{eqn-2}
  \bm{t}_{s}^{j} =\left [\omega_{1}, \omega_{2}, \cdots, \omega_{M}  \right ]~\text{,} 
\end{equation}
where $M$ denotes the count of learnable tokens that is set equal to $L$+1.
$\bm{t}_{s}^{j}$ excludes class word embedding to assist $\bm{t}_{c}$ capture context and reduce class-specific bias.
To achieve classification outcomes, we utilize image $\bm{x}$ and prompt pair $\bm{t} = \{\bm{t}_c,\bm{t}_{s}\} $, 
which are input into the respective encoders $\mathcal{F}(\cdot)$ and $\mathcal{G}(\cdot)$ to extract distinctive features: $ \bm{f}_{x} = \mathcal{F}(\bm{x})$, $\bm{g}_{c} = \mathcal{G}(\bm{t}_{c})$ and $ \bm{g}_{s}=\mathcal{G}(\bm{t}_{s})$.
The final predicted score is formulated as follows: 
\begin{align}\label{eqn-3} 
  \bm{p}=\text{CFA}(\bm{f}_{x}, \bm{g}_{c}) \odot \bm{g}_{c} -\text{CFA}(\bm{f}_{x}, \bm{g}_{c}) \odot \bm{g}_{s}~\text{,}
\end{align}
where $\odot$ symbolizes Hadamard product, $\text{CFA}(\cdot, \cdot)$ denotes the class-specific region feature aggregation function~\cite{sun2022dualcoop}  and its calculation process is as follows:
\begin{equation}\label{eqn-4} 
  \begin{split}
    & \bm{f}_{i\to t }=\text{Proj}_{i \to t } \left ( \bm{f}_{x} \right ) \\
    & \bm{f}_{c}=\text{softmax}(\bm{f}_{i\to t }\cdot \bm{g}_{c}^{\top})\cdot \bm{f}_{i\to t }~\text{,}
    % & F_{s}=softmax(F_{i\to t }\cdot T_{s}^{\top})\cdot F_{i\to t }.
  \end{split}
\end{equation}
where $\text{Proj}_{i \to t }(,)$ projects the image features into the textual feature space. Within 
$\bm{f}_{c}\in \mathbb{R}^{n\times d}$, each 
vector $\bm{f}_{c}^{i}\in \mathbb{R}^{1 \times d}$ specifically encapsulates the features related to category \textit{i}. 
\vspace{-1.0em}
\subsection{Selective Confidence Cluster Replay}
Recent rehearsal-based methods~\cite{dong2023knowledge,rebuffi2017icarl}  rely on using average features for category representation, resulting in blurred distinct category features and obscure less salient targets in MLCIL.
To overcome this limitation, we introduce the Selective Confidence Cluster Replay (SCCR) strategy.

Initially, we employ Eq.~\eqref{eqn-4} to distill category-related features $\bm{f}$ from each individual sample. Here, we set $\bm{f}=\bm{f}_{c}$, because we primarily sample based on the ROI of positive samples.
% Taking into account intra-class variation within the same category, 
We then employ the K-means clustering algorithm to ensure sample diversity by partitioning the feature set of each category into \textit{m} distinct clusters.
To fully leverage the limited old samples and enhance the model's robustness, we further consider sample selection based on the model's performance.
To this end, we introduce a confidence-based cluster sampling approach.
In detail, within each cluster, we select 
\textit{k} hard samples for retraining.
 The sample set of samples for each category can be obtained by the following process(omitting category identification for brevity):
 \begin{equation}\label{eqn-5} 
  \mathcal{R}_{i} = \text{Top}_{k}^{low}  \left ( \left \{ \bm{x}_{j}\mid \bm{x}_{j}\in \bm{G}_{i},p_{j}(\bm{x}_{j}\mid y_{j},\theta _{s-1})  \right \}  \right )~\text{,}
\end{equation}
where $\mathcal{R}_{i}$ denotes the set of samples selected from cluster $\bm{G}_{i}$
 , with $\bm{x}_{j}$ representing each sample and $p_{j}$ representing the predicted probability of $\bm{x}_{j}$
  by model $\theta_{\mathrm{s-1}}$. The $\text{Top}_{k}^{low}$ identifies the \textit{k} samples with the lowest $p_{j}$ values from $\bm{G}_{i}$.
  The final set of samples is $\mathcal{R}=\mathcal{R}_{1} \cup \mathcal{R}_{2} \cdots \cup \mathcal{R}_{m}$.
 Then $\mathcal{R}$ is combined with the training set $\mathcal{D}_{tr}^s$, facilitating the training process in session 
\textit{s}: $\widetilde{\mathcal{D}}_{tr}^s =\mathcal{D}_{tr}^s\cup \mathcal{R}$.
\vspace{-1.0em}
\subsection{Optimization Objective}
In this study, we utilize Asymmetric Loss~\cite{ridnik2021asymmetric} (ASL) that is widely used in multi-label classification tasks, to optimize the parameters of the prompts.
\begin{equation}\label{eqn-6} 
  \mathcal{L}_{asl}=\frac{1}{M} \sum_{m=1}^{M}\left\{\begin{array}{ll}
    \left(1-p_{m}\right)^{\gamma+} \log \left(p_{m}\right), & \text { if } y_{m}=1 \\
    \left(p_{m}\right)^{\gamma-} \log \left(1-p_{m}\right), & \text { if } y_{m}=0
    \end{array}\right.~\text{,}
\end{equation}
where $y_{m}$ is a binary label that signifies the presence of category $m$ within the sample. Additionally, $\gamma+$ and $\gamma-$ serve as the attention weights for positive and negative samples. 
Furthermore, we introduce $\mathcal{L}_{tpc}$ to ensure the consistency of the prompt.
The computation process is as follows:
\begin{equation}\label{eqn-7} 
  % \mathcal{L}_{tpc} =2-\left \langle T_{c}^{s},T_{c}^{s-1}  \right \rangle -\left \langle T_{s}^{s},T_{s}^{s-1}  \right \rangle~\text{.}
  \mathcal{L}_{tpc} =2 - \text{cos}(\bm{g}_{c}^{s}, \bm{g}_{c}^{s-1}) - \text{cos}(\bm{g}_{s}^{s}, \bm{g}_{s}^{s-1} )~\text{,}
\end{equation}
where $\text{cosine}(\cdot,\cdot)$ denotes cosine similarity. 
$\bm{g}_{type\in \left \{c, s  \right \}}^{s}$ and  $\bm{g}_{type\in \left \{c, s \right \}}^{s-1}$ represent the textual features of the old categories extracted by models 
$\theta _{s}$ and $\theta _{s-1}$  respectively.
Finally, the overall optimization objective of our model is: $\mathcal{L}= \mathcal{L}_{asl} +\alpha \mathcal{L}_{tpc}~\text{,}$
where $\alpha$ is balancing factor.
\vspace{-1.0em}
\section{Evaluation}
\subsection{Experiment setup}
\textbf{Datasets and Metrics.} We conduct experiments on the \textbf{MS-COCO} \cite{lin2014microsoft} and \textbf{PASCAL VOC} 2007~\cite{everingham2010pascal} datasets 
to evaluate the effectiveness of our approach.
MS-COCO is annotated with 80 distinct categories.  
PASCAL VOC comprises 20 categories. 
We adopt the B\textit{i}C\textit{j}~\cite{dong2023knowledge,wlargeu2019} setup for each dataset, where $i$ denotes the number of classes in the initial training session, and $j$ represents the classes incorporated in each incremental session. 
All categories are arranged alphabetically. We use two key metrics in MLCIL: average accuracy and last accuracy. Average accuracy is the mean mAP score across all sessions, while last accuracy is the mAP score from the final session. We also report per-class F1 (CF1) and overall F1 (OF1) measures.

\noindent \textbf{Implementation Details.} We leverage the pre-trained CLIP model, specifically its ViT-B/16 variant as backbone. The number of learnable prompt embeddings L is set to 16. Images are resized to $224 \times 224$. The model is trained for 20 epochs with batch size of 64 in each session. Optimization is performed using the Adam~\cite{kingma2014adam} and the OneCycleLR scheduler with 
a weight decay of 1e-4. During the incremental session, learning rates are set to 2e-4 for MS-COCO and 1.6e-3 for PASCAL VOC. Moreover, base sessions for all experiments initiate with learning rate of 1.6e-3.
We report the upper bounds of model performance: Joint. Joint denotes the results obtained 
by training the model using all categories from the entire dataset within single session.
\vspace{-2.0em}
\begin{center}
\begin{table*}[ht]
  \renewcommand\arraystretch{1.26}
  \footnotesize
  \centering
  \setlength{\abovecaptionskip}{0cm}
  \caption{Experimental results on MS COCO dataset. 
  The best results are shown in \textbf{bold}.
  "*" represents the results reported in~\cite{dong2023knowledge}, and the same applies to the following.}
  \label{table1}
  \resizebox{\textwidth}{!}{
  \begin{tabular}{cccccccccc}
    \toprule[1pt]
    \multicolumn{1}{c}{\multirow{3}{*}{Methods}} & \multicolumn{1}{c}{\multirow{3}{*}{Buffer Size}} & \multicolumn{4}{|c|}{MS-COCO B0-C10} & \multicolumn{4}{c}{MS-COCO B40-C10} \\
    \cline{3-10}
    \multicolumn{1}{c}{} & & \multicolumn{1}{|c|}{Avg.Acc} & \multicolumn{3}{c}{Last Acc} & \multicolumn{1}{|c|}{Avg.Acc} & \multicolumn{3}{c}{Last Acc} \\
    \cline{3-10}
    \multicolumn{1}{c}{} & & \multicolumn{1}{|c|}{mAP(\%)} & \multicolumn{1}{c}{CF1} & \multicolumn{1}{c}{OF1} & \multicolumn{1}{c}{mAP(\%)} & \multicolumn{1}{|c|}{mAP(\%)} & \multicolumn{1}{c}{CF1} & \multicolumn{1}{c}{OF1} & \multicolumn{1}{c}{mAP(\%)} \\
    \midrule
    \multicolumn{1}{c|}{Joint} & - & \multicolumn{1}{|c|}{-} & 77.8 & 81.2 & 83.9 & \multicolumn{1}{|c|}{-} & 77.8 & 81.2 & 83.9 \\
    \midrule  
    \multicolumn{1}{c|}{Lwf$^{*}$~\cite{li2017learning}} & \multicolumn{1}{c}{\multirow{4}{*}{0}} & \multicolumn{1}{|c|}{47.9} & 9.0 & 15.1 & 28.9 & \multicolumn{1}{|c|}{48.6} & 9.5 & 15.8 & 29.9 \\
    \multicolumn{1}{c|}{KRT~\cite{dong2023knowledge}} & & \multicolumn{1}{|c|}{74.6} & 55.6 & 56.5 & 65.9 & \multicolumn{1}{|c|}{77.8} & 64.4 & 63.4 & 74.0 \\
     \multicolumn{1}{c|}{MULTI-LANE~\cite{de2024less}} & & \multicolumn{1}{|c|}{79.1} & 65.1 & 62.8 & \textbf{74.5} & \multicolumn{1}{|c|}{78.8} & 66.0 & 66.6 & 76.6 \\
    \multicolumn{1}{c|}{Ours} & & \multicolumn{1}{|c|}{\textbf{80.7}} & 65.1 & \textbf{65.1} & 73.3 & \multicolumn{1}{|c|}{\textbf{81.1}} & \textbf{68.2} & \textbf{68.5} & \textbf{77.2} \\
    \midrule
    \multicolumn{1}{c|}{iCaRL~\cite{rebuffi2017icarl}} & \multicolumn{1}{c}{\multirow{6}{*}{20/class}} & \multicolumn{1}{|c|}{59.7} & 19.3 & 22.8 & 43.8 & \multicolumn{1}{|c|}{65.6} & 22.1 & 25.5 & 55.7 \\
    \multicolumn{1}{c|}{ER~\cite{riemer2018learning}} & & \multicolumn{1}{|c|}{60.3} & 40.6 & 43.6 & 47.2 & \multicolumn{1}{|c|}{68.9} & 58.6 & 61.1 & 61.6 \\
    \multicolumn{1}{c|}{Der++~\cite{buzzega2dark020}} & & \multicolumn{1}{|c|}{72.7} & 45.2 & 48.7 & 58.8 & \multicolumn{1}{|c|}{71.0} & 46.6 & 42.1 & 64.2 \\
    \multicolumn{1}{c|}{AGCN-R~\cite{du2023multi}} & & \multicolumn{1}{|c|}{73.2} & 59.5 & 60.3 & 66.0 & \multicolumn{1}{|c|}{75.2} & 64.1 & 65.2 & 71.7 \\
    \multicolumn{1}{c|}{KRT-R$^{*}$~\cite{dong2023knowledge}} & & \multicolumn{1}{|c|}{76.5} & 63.9 & 64.7 & 70.2 & \multicolumn{1}{|c|}{78.3} & 67.9 & 68.9 & 75.2 \\
    \multicolumn{1}{c|}{Ours} & & \multicolumn{1}{|c|}{\textbf{81.7}} & \textbf{69.6} & \textbf{72.0} & \textbf{76.1} & \multicolumn{1}{|c|}{\textbf{81.5}} & \textbf{71.8} & \textbf{74.7} & \textbf{78.4} \\
    \midrule
    \multicolumn{1}{c|}{OCDM~\cite{liang2022optimizing}} & \multicolumn{1}{c}{\multirow{3}{*}{1000}} & \multicolumn{1}{|c|}{49.5} & 9.3 & 14.9 & 28.7 & \multicolumn{1}{|c|}{51.5} & 10.0 & 16.2 & 34.9 \\
    \multicolumn{1}{c|}{KRT-R~\cite{dong2023knowledge}} & & \multicolumn{1}{|c|}{75.7} & 61.6 & 63.6 & 69.3 & \multicolumn{1}{|c|}{78.3} & 67.5 & 68.5 & 75.1 \\
    \multicolumn{1}{c|}{Ours} & & \multicolumn{1}{|c|}{\textbf{81.9}} & \textbf{69.6} & \textbf{71.8} & \textbf{78.5} & \multicolumn{1}{|c|}{\textbf{81.6}} & \textbf{71.7} & \textbf{74.5} & \textbf{78.4} \\
    \bottomrule[1pt]
  \end{tabular}
     }    
\end{table*}  
\end{center}
\vspace{-5.0em}
\subsection{Comparsion results}
Table~\ref{table1} and Table~\ref{table20} summarize the results of our experiments conducted on the MS-COCO and PASCAL VOC datasets. 
Notably, the "Buffer Size" column represents the quantity of samples retained for replay.

\noindent \textbf{Results on MS-COCO.} Table~\ref{table1} presents the experimental results of our method compared with other approaches under B40-C10 and B0-C10. Firstly, when training is conducted without rehearsal buffer, our average accuracy surpasses MULTI-LANE~\cite{de2024less} by \textbf{1.6\%}, and achieves last accuracy of \textbf{73.3\%}. 
When the buffer size is set to 20 per class, our model achieve last accuracy of \textbf{78.4\%} under B40-C10, 
which is \textbf{3.2\%} higher than the second-best method.
Notably, when the buffer size reaches 1000, our method outperforms other methods,
with final accuracy that is \textbf{3.3\%} higher than KRT~\cite{dong2023knowledge}.
This indicates that our \textit{SCCR} module demonstrates superior performance in effective sample acquisition. Table~\ref{table3} reveal that our method's performance discrepancy relative to Joint is \textbf{5.5\%}, positioning it closer to Joint's outcome than the KRT. This underscores our method's enhanced capability to mitigate forgetting.
\begin{center}
% \vspace{-1.0em}
  \begin{table*}[htb]
    \renewcommand\arraystretch{1.26}
    \footnotesize
    \centering
    \caption{Parameter quantity and performance gap with respect to the Joint on MS-COCO dataset under B40-C10 setting. "()" represents the performance gap.}
    \label{table3}
    \resizebox{0.7\textwidth}{!}{
    \begin{tabular}{c|c|c|cc}
      \toprule[1pt]
      Methods & Backbone & Param. & Avg.mAp($\%$) & Last.mAP($\%$) \\
      \midrule
      Joint & \multicolumn{1}{c|}{\multirow{2}{*}{TResNetM}} & \multicolumn{1}{c|}{\multirow{2}{*}{29.4M}} & - & 81.8 \\
      KRT-R~\cite{dong2023knowledge} & & & 78.3 & 75.2(\textbf{$\downarrow$6.6}) \\
      \midrule
      Joint & \multicolumn{1}{c|}{\multirow{2}{*}{ViT-B/16}} & \multicolumn{1}{c|}{\multirow{2}{*}{28.5M}} & - & 83.9 \\
      Ours & & & 81.5 & 78.4(\textbf{$\downarrow$5.5}) \\
      \bottomrule[1pt]
    \end{tabular}
      }
      % \vspace{-2.0em} 
  \end{table*}  
  \end{center}
\begin{center}
  \begin{table*}[ht]
   \vspace{-1.0em}
    \renewcommand\arraystretch{1.26}
    \footnotesize
    \centering
    \caption{Experimental results on PASCAL VOC dataset.  }
    \label{table20}
    \resizebox{0.8\textwidth}{!}{
    \begin{tabular}{cccccccccc} 
      \toprule[1pt]
      \multicolumn{1}{c}{\multirow{2}{*}{Methods}} & \multicolumn{1}{c}{\multirow{2}{*}{Buffer Size}} & \multicolumn{2}{|c|}{VOC B0-C4} & \multicolumn{2}{c}{VOC B10-C2} \\
      \cline{3-6}
      \multicolumn{1}{c}{} & & \multicolumn{1}{|c}{Avg.Acc} & \multicolumn{1}{c|}{Last Acc} & \multicolumn{1}{c}{Avg.Acc} & \multicolumn{1}{c}{Last Acc} \\
      \midrule
      \multicolumn{1}{c}{Joint} & \multicolumn{1}{|c|}{-} & - & \multicolumn{1}{c|}{94.41} & - & 94.41 \\
      \midrule
      \multicolumn{1}{c|}{AGCN~\cite{du2023multi}} & \multicolumn{1}{c}{\multirow{5}{*}{0}} & \multicolumn{1}{|c}{84.3} & \multicolumn{1}{c|}{73.4} & 79.4 & 65.1 \\
       \multicolumn{1}{c|}{KRT~\cite{dong2023knowledge}} & & \multicolumn{1}{|c}{89.5} & \multicolumn{1}{c|}{74.8} & 82.9 & 67.9 \\
        \multicolumn{1}{c|}{MULTI-LANE~\cite{de2024less}} & & \multicolumn{1}{|c}{\textbf{93.5}} & \multicolumn{1}{c|}{\textbf{88.8}} & \textbf{93.1} & \textbf{88.3} \\
      \multicolumn{1}{c|}{Ours} & & \multicolumn{1}{|c}{\textbf{93.5}} & \multicolumn{1}{c|}{88.1} & 88.5 & 79.0 \\
      \midrule
      \multicolumn{1}{c|}{TPCIL~\cite{tao2020topology}} & \multicolumn{1}{c}{\multirow{6}{*}{2/class}} & \multicolumn{1}{|c}{87.6} & \multicolumn{1}{c|}{77.3} & 80.7 & 70.8 \\
      \multicolumn{1}{c|}{PODNet~\cite{douillard2020podnet}} & & \multicolumn{1}{|c}{88.1} & \multicolumn{1}{c|}{76.6} & 81.2 & 71.4 \\
      \multicolumn{1}{c|}{Der++~\cite{buzzega2dark020}} & & \multicolumn{1}{|c}{87.9} & \multicolumn{1}{c|}{76.1} & 82.3 & 70.6 \\
      \multicolumn{1}{c|}{AGCN~\cite{du2023multi}} & & \multicolumn{1}{|c}{86.5} & \multicolumn{1}{c|}{76.0} & 82.8 & 69.3 \\
      \multicolumn{1}{c|}{KRT~\cite{dong2023knowledge}} & & \multicolumn{1}{|c}{90.7} & \multicolumn{1}{c|}{83.4} & 87.7 & 80.5 \\
      \multicolumn{1}{c|}{Ours} & & \multicolumn{1}{|c}{\textbf{93.8}}
      & \multicolumn{1}{c|}{\textbf{88.1}} & \textbf{90.9} & \textbf{84.1} \\
      \bottomrule[1pt]
    \end{tabular}
      }
  \vspace{-2.0em}    
  \end{table*}  
  \end{center}
\noindent \textbf{Results on PASCAL VOC.} Table~\ref{table20} presents the results of our method compared to other strategies on the PASCAL VOC dataset under B10-C2 and B0-C4 settings. 
Specifically, with a buffer size of 0, we achieve best average precision under B0-C4. With a buffer size of 2 samples per class, the mean accuracy increases from \textbf{88.5\%} to \textbf{90.9\%} under the B10-C2 scenario. In the final session of the B10-C2 setup, our method leads the second-best by \textbf{3.6\%}. These results indicate strong performance and potential to approach upper bound performance.
\subsection{Qualitative results}
\begin{figure}[ht]
\centering
% \vspace{-2.0em}
\includegraphics[width=1.0\textwidth]{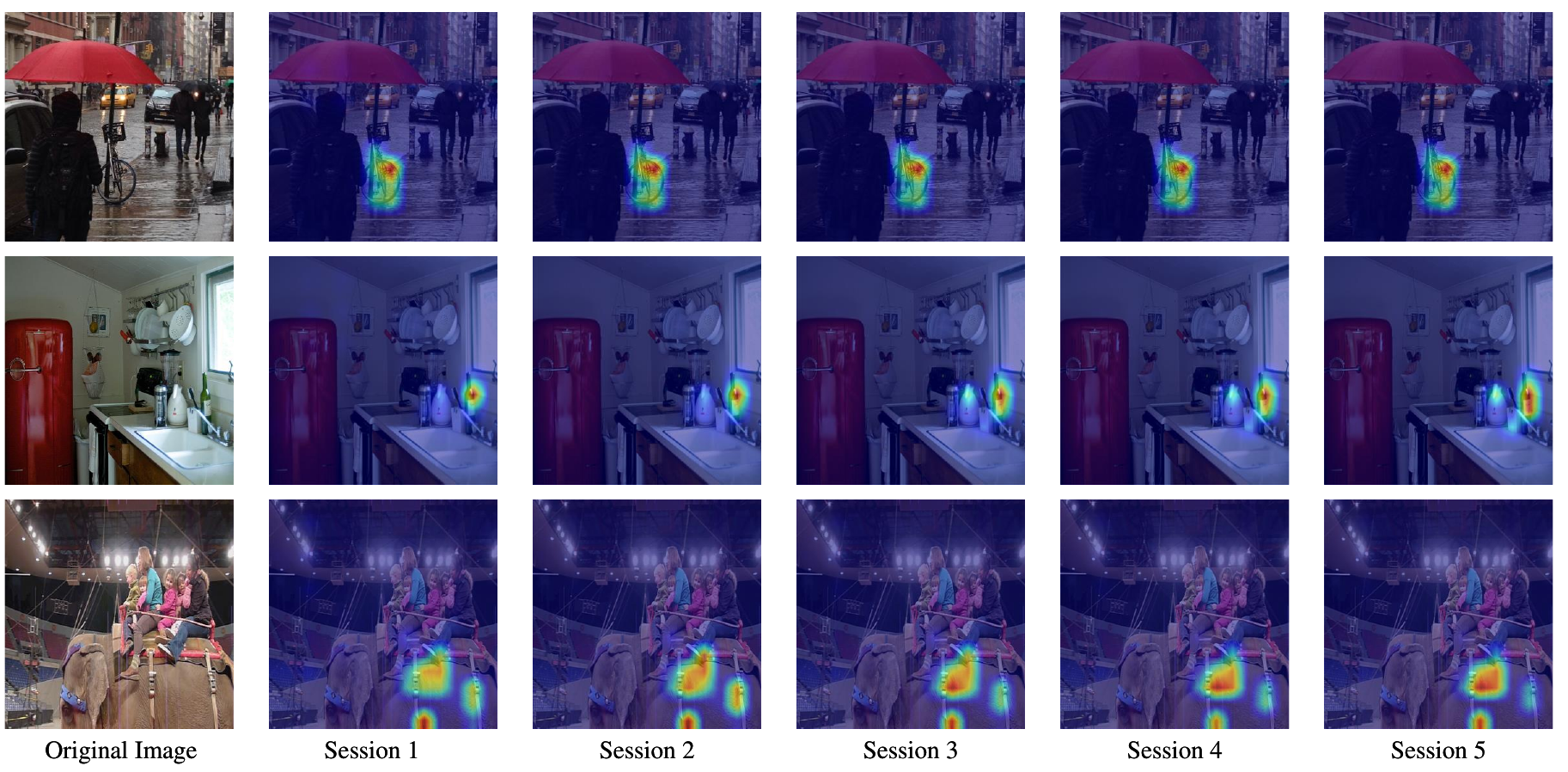}
\vspace{-2.0em}
\caption{Visualization of the attention maps on MS-COCO under B40-C10 setting.}
\label{figure3}
\end{figure}

In incremental learning, attention region maps indicate what the model retains or discards. Fig.~\ref{figure3} illustrates the evolution of these maps across training sessions, showing consistent focus on base categories \text{\ttfamily{\{bicycle;bottle;elephant\}}}. As training progresses, the attention regions for each category remain consistent, suggesting the model effectively retains knowledge of existing categories while integrating new information. 
\vspace{-3.0em}
\begin{table*}[ht]
\begin{center}
    \renewcommand\arraystretch{1.26}
    \footnotesize
    \centering
    \setlength{\abovecaptionskip}{0cm}
    \caption{The results of the ablation study on the effectiveness of different components.}
    \label{table4}
     \resizebox{0.7\textwidth}{!}{
    \begin{tabular}{c|ccc|cc}
      \toprule[1pt]
      Model & \text{ ICP } & \text{ SCCR } & \text{ } $\mathcal{L}_{tpc} \text{ }$ & \text{ } Avg.Acc \text{ } & \text{ } Last Acc \text{ } \\
      \midrule
      Baseline & \no & \no & \no & 79.7 & 75.9(+\textbf{0.0})  \\
      \midrule
      (1) w/ ICP & \yes & \no & \no & 80.8 & 76.7(+\textbf{0.8}) \\
      \quad(2) w/ SCCR & \no & \yes & \no & 80.6 & 77.0(+\textbf{1.1}) \\
      \enspace(3) w/o $\mathcal{L}_{tpc}$ & \yes & \yes & \no & 81.1 & 77.9(+\textbf{2.0}) \\
      \quad \enspace(4) w/o SCCR & \yes & \no & \yes & 81.1 & 77.2(+\textbf{1.3}) \\
      \midrule
     Ours & \yes & \yes & \yes & 81.5 & 78.4(+\textbf{2.5})  \\
      \bottomrule[1pt]
    \end{tabular}
       }
      \vspace{-2.0em} 
\end{center}
\end{table*}  
\subsection{Ablation Studies}
Table~\ref{table4}  presents the results of the ablation study on MS-COCO under B40-C10 setting, with a buffer size of 20 per class. We integrate the CoOp with the CFA($\cdot$,$\cdot$) function and concurrently employ DPL~\cite{dong2023knowledge} for anti-forgetfulness as the
baseline approach. The results indicate that both ICP and SCCR significantly improve performance compared to the baseline. SCCR alone improves last accuracy by \textbf{1.1\%}, while adding ICP increases this to \textbf{2.0\%}, demonstrating the strong anti-forgetting capabilities of both components.
Additionally, we observe that incorporating $\mathcal{L}_{tpc}$ further boosts performance, with \textbf{1.3\%} increase in last accuracy when combined with ICP. Ultimately, combining all three components yields \textbf{2.5\%} improvement over the baseline, demonstrating the effectiveness of our approach in MLCIL.
\section{Conclusion}
% \vspace{-0.5em}
In summary, we introduce a novel method for addressing Multi-Label Class-Incremental Learning (MLCIL) by integrating Incremental Context Prompting (ICP) and Selective Confidence Cluster Replay (SCCR). ICP learns a pair of prompts for each category to enhance the model’s focus on all categories within an image, while SCCR uses confidence-based sampling to ensure sample diversity and efficient use of limited samples. Our approach advances the use of image-text matching in MLCIL. Experimental results on the MS COCO and PASCAL VOC datasets demonstrate its effectiveness and competitive performance.
% \vspace{-1.0em}
\section*{Acknowledgements}\small{
This work was supported in part by the 2023 New Era Education Quality Project (Postgraduate Education, No.20231hpysfjd009), the National Natural Science Foundation of China (No.62472004), Natural Science Foundation of Anhui Province (No.2308085MF214), University Synergy Innovation Program of Anhui Province (No.GXXT-2022-029), Key Natural Science Project of Anhui Provincial Education Department (No.2023AH050065).
We also thank the High-performance Computing Platform of Anhui University for providing computational resources for this project.
}
\bibliographystyle{splncs04}
\bibliography{2024.bib}
\clearpage
\end{document}